\documentclass[a4paper,fleqn]{cas-dc}

\usepackage[authoryear,longnamesfirst]{natbib}

\def\tsc#1{\csdef{#1}{\textsc{\lowercase{#1}}\xspace}}
\tsc{WGM}
\tsc{QE}
\tsc{EP}
\tsc{PMS}
\tsc{BEC}
\tsc{DE}

\usepackage{amsmath,amsfonts}
\usepackage{algorithmic}
\usepackage{array}
\usepackage{textcomp}
\usepackage{stfloats}
\usepackage{url}
\usepackage{verbatim}
\usepackage{graphicx}

\usepackage{algorithm}
\usepackage{comment}
\usepackage{amssymb}
\usepackage{amsthm}
\usepackage{dsfont}
\usepackage{xcolor}

\begin{document}
\let\WriteBookmarks\relax
\def\floatpagepagefraction{1}
\def\textpagefraction{.001}

\shorttitle{LAECIPS}


\title [mode = title]{LAECIPS: Large Vision Model Assisted Adaptive Edge-Cloud Collaboration for IoT-based Embodied Intelligence System}                      



%
\author[inst1]{Shijing Hu}

\author[inst1]{Zhihui Lu}
\author[inst2]{Xin Xu}

\author[inst1]{Ruijun Deng}

\author[inst1]{Xin Du}

\author[inst3]{Qiang Duan}

\affiliation[inst1]{organization={School of Computer Science},
            addressline={Fudan University}, 
            city={Shanghai},
            country={China}}

\affiliation[inst2]{organization={School of Information Management},
            addressline={Shanghai Lixin University of Accounting and Finance}, 
            city={Shanghai},
            country={China}}

\affiliation[inst3]{organization={Information Sciences and Technology Department},
            addressline={Pennsylvania State University}, 
            city={Abington},
            country={USA}}



\begin{abstract}
Embodied intelligence (EI) enables manufacturing systems to flexibly perceive, reason, adapt, and operate within dynamic shop floor environments. In smart manufacturing, a representative EI scenario is \textbf{robotic visual inspection}, where industrial robots must accurately inspect components on rapidly changing, heterogeneous production lines. This task requires both high inference accuracy—especially for uncommon defects—and low latency to match production speeds, despite evolving lighting, part geometries, and surface conditions. To meet these needs, we propose \textbf{LAECIPS}, a large vision model-assisted adaptive edge-cloud collaboration framework for IoT-based embodied intelligence systems. LAECIPS decouples large vision models in the cloud from lightweight models on the edge, enabling plug-and-play model adaptation and continual learning. Through a hard input mining-based inference strategy, LAECIPS routes complex and uncertain inspection cases to the cloud while handling routine tasks at the edge, achieving both high accuracy and low latency. Experiments conducted on a real-world robotic semantic segmentation system for visual inspection demonstrate significant improvements in accuracy, processing latency, and communication overhead compared to state-of-the-art methods. LAECIPS provides a practical and scalable foundation for embodied intelligence in smart manufacturing, especially in adaptive robotic inspection and quality control scenarios.

\end{abstract}




\maketitle

\section{Introduction} \label{introdcution}

The rapid advancement of Machine Learning (ML) has greatly empowered intelligent perception in Internet of Things (IoT) applications such as robotic visual inspection, industrial automation, and autonomous driving~\cite{prakash2021multi}. In the context of robotics and smart manufacturing, IoT-based perception systems play a key role in enabling \textit{embodied intelligence} (EI)~\cite{long2023human}—the ability of physical systems, such as robots and manufacturing equipment, to perceive, interpret, reason, and adapt within complex production environments~\cite{liu2025embodied}. These intelligent systems increasingly rely on high-accuracy, low-latency ML inference to support tasks like quality inspection, process monitoring, and flexible assembly~\cite{zhou2019edge}. Typically, ML models are deployed at the edge, close to equipment and operators, to reduce latency. However, resource constraints on edge devices limit their ability to run advanced, high-capacity models~\cite{shuvo2022efficient}, while lightweight models may suffer poor inference accuracy, particularly on novel or rare cases~\cite{zhang2022advancing}.
Further, dynamic and unpredictable environmental changes (such as new product variants, equipment wear, or customized processes) in smart manufacturing can lead to \textit{data distribution drifts}~\cite{de2021continual}. This often renders static, pre-trained models suboptimal for new conditions, thus demanding a framework that supports continual learning and adaptation~\cite{zhou2025decentralized}.

\begin{figure}[t]
\centering
\includegraphics[width=1.0\columnwidth]{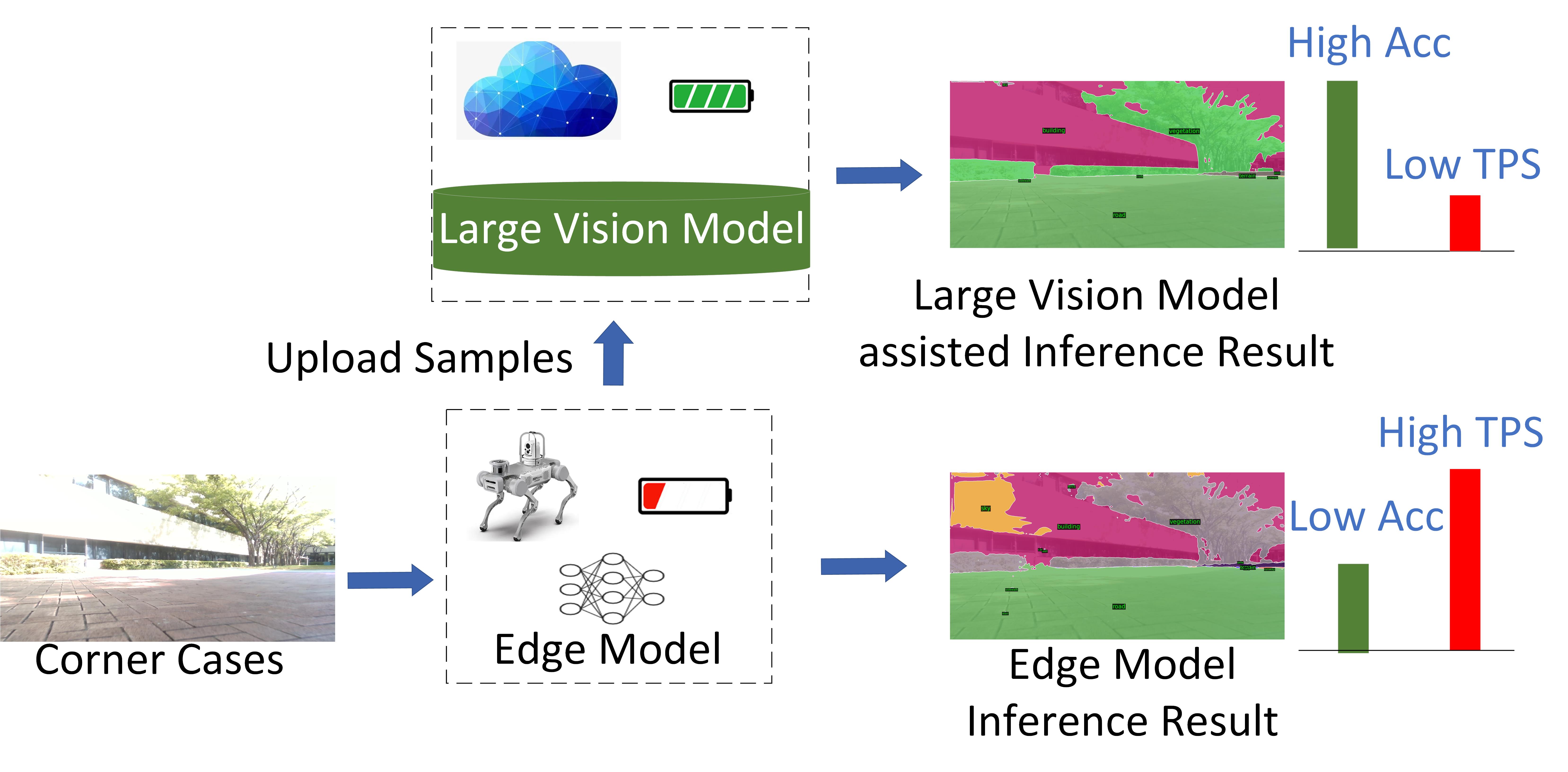} 
\caption{Scenario of Edge Cloud Collaborative Inference for IoT-based Embodied Intelligence System in Robotic Visual Inspection Applications}
\label{senario}
\end{figure}

A prominent and practically significant scenario in embodied intelligence for smart manufacturing is \textbf{robotic visual inspection}~\cite{fan2025embodied}. In real-world factories, vision-guided robots equipped with IoT sensors are deployed for automated inspection of products and components on fast-paced, variable production lines~\cite{ma2024spdp}. Such settings are characterized by dynamic changes: new product models, varying lighting conditions, surface textures, occasional appearance of rare defects, and rapid reconfiguration of inspection tasks~\cite{valenzuela2024embodying}. Achieving consistently high-quality inspection outcomes requires inspection robots to possess embodied intelligence capabilities: robust perception, real-time cognition, adaptive decision-making, and continual self-improvement—all under stringent time constraints~\cite{stella2025synergy}.

Recently, considerable progress has been made in developing large vision models, for example, the Segment Anything Model (SAM) from Meta \cite{kirillov2023segment}. With their strong generalization ability, such large vision models may achieve very high accuracy in handling corner cases and being robust to data distribution drifts in intelligent perception \cite{wssam}. However, a large vision model can only be deployed in a resource-rich cloud data center, which may cause long latency due to data transmissions between user devices and the cloud server. Therefore, how to fully leverage the advantages of a large vision model for achieving accurate inference while reducing perception latency in the resource-constrained IoT becomes an important research problem. 

To answer it, one may think of using edge-cloud collaboration for large-small model co-inference \cite{wang2020convergence, duan2022distributed}. In particular, with a large vision model hosted on the cloud and a small model deployed at the edge, an edge-cloud collaboration strategy determines for each received input if the inference can be performed by the small edge model or needs to be processed by the large vision model on the cloud, as illustrated by Fig.~\ref{senario}. However, existing edge-cloud collaboration methods mainly suffer from three limitations that need to be overcome to support IoT-based intelligent perception. First, the tight coupling between the large and small models limits the system flexibility
of the current methods for fully leveraging large vision models. Second, the collaboration strategy needs to be further optimized for both high accuracy and low latency while demonstrating its capability to adapt to the dynamic IoT environment. Third, inference outputs from a large vision model (e.g., SAM) may lack semantic labels and thus need to be combined with the edge model inference results.  

To address the three issues, we propose LAECIPS, a large vision model-assisted adaptive cloud-edge collaboration framework. In this framework, the large vision model on the cloud and the lightweight model deployed at the edge cooperate in a loose-coupling manner thus making both models plug-and-play and greatly improving the system flexibility. This framework employs an edge-cloud collaboration strategy that is based on hard input mining and optimized for both low response latency and high inference accuracy. This framework also enables online adjustment of the collaboration strategy and continual training of the edge model under the supervision of the large vision model, which makes the system adaptive to data distribution drifts in dynamic IoT environments. 

Specific contributions of this paper are as follows.

\begin{enumerate}
    \item This is the first study that explores the problem of edge-cloud collaborative perception for dynamic IoT data streams in embodied intelligence systems. We propose a novel edge-cloud collaboration framework, LAECIPS, to enable flexible utilization of both large and small models in an online manner to solve this problem.
    
    \item In the LAECIPS framework, we design a hard input mining-based edge-cloud co-inference strategy that achieves higher accuracy and lower task processing latency.
     
    \item In the LAECIPS framework, we propose the continual training of the small model to fit in with the dynamic environmental changes in the IoT-based embodied intelligence environment.
    
    \item We  analyze the 
    theoretical generalization capability of LAECIPS to prove the feasibility of incorporating large vision models, edge small models, and edge-cloud co-inference strategies into the LAECIPS framework in a plug-and-play manner.
    
    \item We implement the proposed LAECIPS framework through a real-world robotic semantic segmentation system in a realistic edge-cloud environment to demonstrate its applicability. Extensive experimental results substantiate that LAECIPS achieves significantly higher accuracy,lower task processing latency and communication overhead than its SoTA competitors.
    
 \end{enumerate}
 
The rest of the paper is structured as follows. Section~\ref{sec: related work} explains the related work. The technical details of LAECIPS are given in Section~\ref{sec:method}. Section~\ref{sec:theory} presents the theoretical proof of the generalization ability of LAECIPS. Experimental results are presented in Section~\ref{sec:evaluation}. Finally, we conclude the paper in Section~\ref{sec: conclusion}.

\section{Related Work}
\label{sec: related work}
The related research on cloud-edge collaborative inference can be categorized into two categories: model partition and big/little model cooperation.

\paragraph{Model Partition}

Model partition segments a (big) model into multiple sub-models that are deployed on different hosts including cloud servers and edge device(s) based on their resource availability. During inference operation, the model is collaboratively computed across all sub-models to obtain the output result.      
For example, Neurosurgeon \cite{kang2017neurosurgeon} uses a performance prediction model to select the optimal split point for a model. JoinDNN \cite{eshratifar2019jointdnn} formulates the optimal model layers scheduling as the shortest path problem and solves it using integer linear programming. DADS \cite{hu2019dynamic} formulates different model partition optimization problems for lightly and heavily loaded conditions. IONN \cite{jeong2018ionn} incrementally builds the model on the server using the arriving model partitions to enable early-stage training. DeepThings \cite{zhao2018deepthings} fuses grids across layers of DNN to construct a fine-grained model partition. 

Although partitioning a complex model across the cloud and edge device(s) reduces computational costs and improves accuracy, it may introduce significant communication overheads for transmitting the intermediate results of the split model, which is often overwhelming to resource-constrained IoTs. Also, the sub-models deployed on the cloud and edge device(s) are tightly coupled thus limiting the flexibility and adaptability of model partitioning to face the dynamic IoT environments. In addition, it is difficult to directly apply the existing model partition methods to the recently developed large vision models due to their highly complex model structures. 

\paragraph{Big/Little Model Cooperation}

The idea of big/little model cooperation is to deploy a lightweight model on the edge device for simple data inference and use a big model on the cloud for handling difficult data. With an appropriate strategy for model selection, big/little model cooperation may achieve high accuracy and low latency with minimized communication overheads. Also, this approach allows loosely coupled models to be deployed on the edge and cloud for more flexibility and adaptability. Therefore, big/little model cooperation offers a promising approach to intelligent perception in the IoT environment.   

Collaborative inference based on big/little model cooperation was first proposed in SM~\cite{CODES15}, where difficult samples were identified using score margin and uploaded to the cloud for inference. In Cachier~\cite{ICDCS17}, the interaction between edge and cloud was modeled as a caching system to minimize inference latency. CeDLD~\cite{FGCS19} applied big/little model collaborative inference to medical image recognition and identified difficult samples based on image similarity. AppealNet~\cite{DAC21} transformed the edge model into a multi-head structure to simultaneously identify difficult samples while outputting the inference results. EdgeCNN~\cite{TCC22} proposed a collaborative training method for big/little model cooperation that uses large vision model outputs to supervise the training of the small model on an edge device. The newly reported SOTA work is DCSB~\cite{cao2023edge}, which applied big/little model collaborative inference to object detection and adaptively down-sampled some regions of the difficult case to reduce bandwidth consumption. Besides that, works that focused on difficult data detection could also bring insights for big/little model collaboration. MESS~\cite{ECCV22} proposed an early exit method for semantic segmentation tasks, which could also be used in detecting difficult data. SPP~\cite{ICLR18} proposed a confidence score-based method to detect out-of-distribution examples which could also be regarded as hard samples.

Although encouraging progress has been made in this area, the SOTA technologies for big/little model cooperation still have some limitations that need to be overcome to effectively support intelligent perception in IoTs. Particularly, the current methods lack the capability of online updating for the edge model and adaptive adjustment of the collaboration strategy in response to the dynamic IoT environments. Also, with the rise of large vision models, current methods need further optimization to apply to large vision models.  

\begin{figure*}[t]
\centering
\includegraphics[width=2.0\columnwidth]{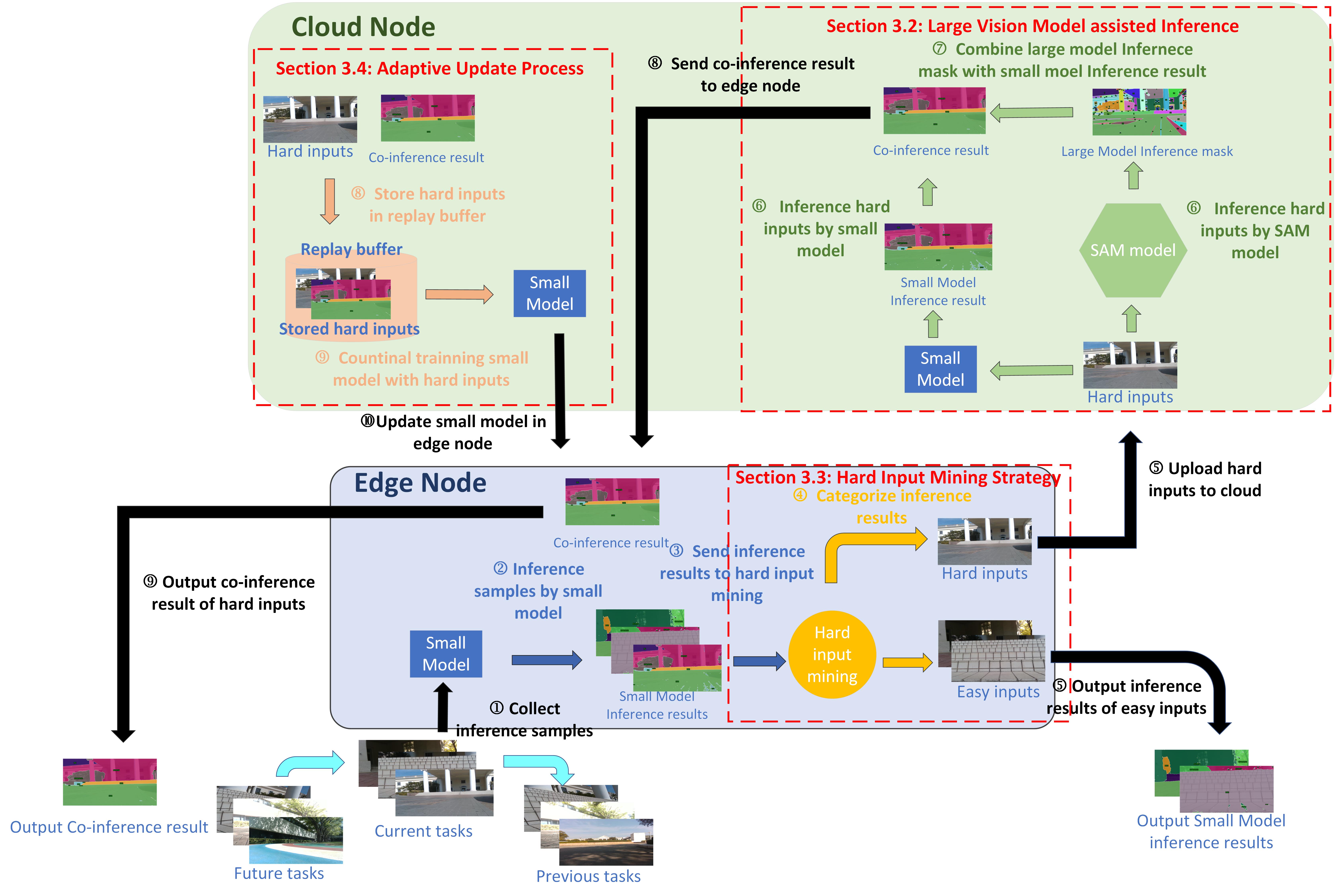} 
\caption{Overview Architecture of LAECIPS}
\label{system-architecture}
\end{figure*}

\section{Proposed Method}
\label{sec:method}

\subsection{System Overview}

We examine a scenario wherein a robot or autonomous vehicle outfitted with a camera is deployed at the edge. The edge node is tasked with executing real-time semantic segmentation duties, while the cloud node acts as a resource-abundant computing center, offering assistance to the edge node. In response to challenges related to the poor performance of edge models when confronted with corner cases, alongside concerns of data drift and heterogeneity within the edge environment, we have devised the LAECIPS architecture, illustrated in Figure~\ref{system-architecture}.

In this framework, a small semantic segmentation model is deployed on the edge device. In steps $\normalsize{\textcircled{\scriptsize{1}}}$ and $\normalsize{\textcircled{\scriptsize{2}}}$, the small model conducts inference on the collected data inputs to yield small model inference results. Subsequently, in steps $\normalsize{\textcircled{\scriptsize{3}}}$ and $\normalsize{\textcircled{\scriptsize{4}}}$, the hard input mining module processes these results to categorize the collected data into two groups: hard inputs and easy inputs. In step $\normalsize{\textcircled{\scriptsize{5}}}$, the small model inference results of easy inputs, which have achieved an acceptable level of accuracy, are directly outputted for reducing processing latency. Conversely, hard inputs that cause low accuracy in edge inference are uploaded to the cloud for further processing to improve inference accuracy. 

In step $\normalsize{\textcircled{\scriptsize{6}}}$, both the small model and the SAM large vision model deployed in the cloud perform their respective inference on the uploaded hard inputs. In step $\normalsize{\textcircled{\scriptsize{7}}}$, a fusion of the cloud inference masks with the small model inference results yields co-inference results. In step $\normalsize{\textcircled{\scriptsize{8}}}$ and $\normalsize{\textcircled{\scriptsize{9}}}$, the cloud node sends the co-inference results to the edge node, which then outputs the co-inference results as the inference results for hard inputs. Additionally, the hard inputs and the co-inference results are stored in the cloud node's replay buffer. In step $\normalsize{\textcircled{\scriptsize{9}}}$, upon the replay buffer's sample count exceeding a predetermined threshold or a specified time interval elapsing, the cloud node proceeds to continually train the small model, using the hard inputs and their co-inference results as the ground truth. Finally, in step $\normalsize{\textcircled{\scriptsize{10}}}$, the small model deployed in the edge node is updated by the small model trained in the cloud node. 

\subsection{Large Vision Model assisted Inference}
\label{3.2}
\begin{figure*}[t]
\centering
\includegraphics[width=2.0\columnwidth]{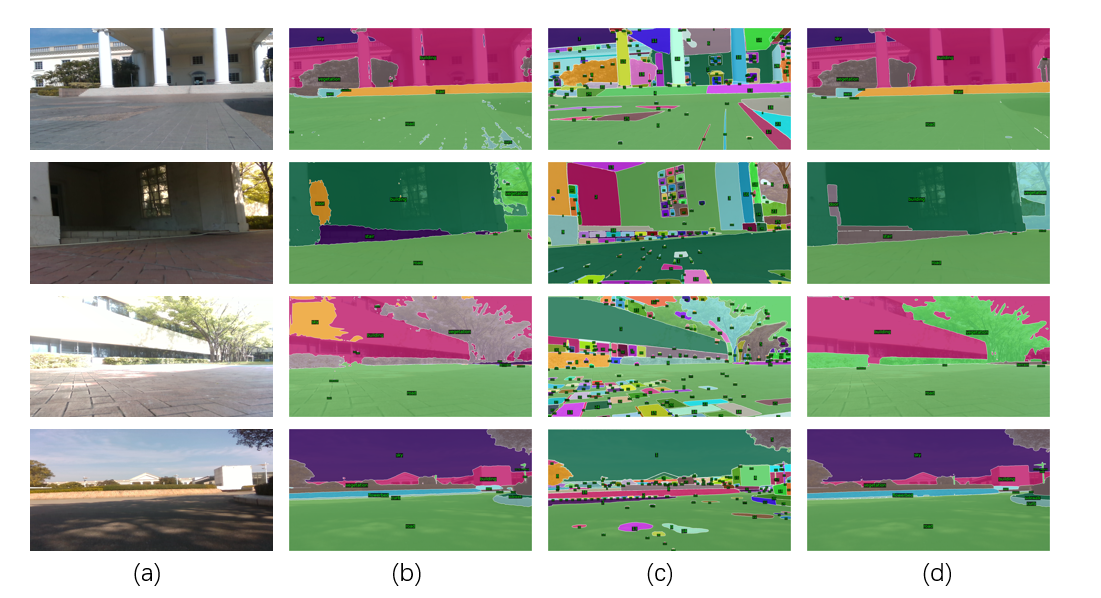} 
\caption{Examples of Inference Samples and their Inference results: (a)Inference Samples, (b) Edge Small Model Inference Results, (c) Cloud SAM Model Inference results, (d) SAM Model assisted Inference Results}
\label{examples}
\end{figure*}

SAM is one of the most representative large vision models for IoT embodied intelligence systems developed in recent years. Its prowess lies in its remarkable efficacy in image segmentation tasks, attributed to its strong generalization ability. Nonetheless, as illustrated in Figure~\ref{examples}c, the SAM model, while adept at producing well-defined contours delineating segmented objects, falls short in providing semantic labels for these segmented images~\cite{wssam}. Consequently, it cannot be directly applied to semantic segmentation tasks. Edge models, in the course of inference, can provide segmented outcomes accompanied by semantic labels. However, these results often bear imperfections, notably exemplified by coarse object edges as depicted in Figure~\ref{examples}b. In semantic segmentation tasks, the process of image segmentation frequently poses greater challenges than the subsequent labeling of the segmented outcomes~\cite{CVPR_ss}. Therefore, a natural idea is to combine the segmentation results from the SAM model with the classification labels from the edge model to obtain a more refined segmentation outcome with classification labels, as depicted in Figure~\ref{examples}d.

We assume that the collected image is $x \in [0,255]^{3 \times H \times W}$ where $H$ is the height of the image and $W$ is the width of the image. The corresponding label of the collected image is $y \in \{0,...M-1\}^{H \times W}$, with $M$ representing the number of classes. The image conforms to the probabilistic distribution $P(x,y)$. Furthermore, we denote the edge model as $f$: $f(x) = y^* \in [0,1]^{M \times H \times W}$, and the large vision model in cloud as $SAM$: $SAM(x) = mask \in \{valid\_mask\}^{ann}$. Notably, the output produced by the $SAM$ model is the mask of image segmentation. During the process of SAM model-assisted semantic segmentation inference, a sample is inferred by the edge model, resulting in a labeled albeit inaccurate segmentation outcome. Concurrently, the same sample is transmitted to the cloud and inferred by the $SAM$ model, generating an unlabeled yet accurate segmentation outcome. The combination of the $SAM$ model's inference outcome and that of the edge model forms the $SAM$ model-assisted inference outcome, defined as follows with its description in Algorithm~\ref{Joint-Inference}.
\begin{equation}
\label{joint}
    F(x) = Assisted\_Inference(f(x), SAM(x))
\end{equation}

\begin{algorithm}[tb]
\caption{Assisted\_Inference algorithm}
\label{Joint-Inference}
\textbf{Input}: Edge Inference Result: $pred$, Large Vision Model Inference Mask: $mask$, class num: $M$\\
\textbf{Output}: large vision model--assisted Inference Result: $semantic\_mask$
\begin{algorithmic}[1] 
\STATE $semantic\_mask \gets pred$
\FOR{$valid\_mask \in mask$}  
    \STATE $scores \gets [0,...,0]$
    \FOR{$i \in [1,...,M]$}
        \STATE $scores[i] \gets \sum_{i=1}^{M}(pred[i][valid\_mask])$
    \ENDFOR
    \STATE $Top\_1\_class \gets \underset{i\in [1,...,M]}{\arg\max} \ scores[i]$
    \STATE $semantic\_mask[valid\_mask] \gets Top\_1\_class$
\ENDFOR
\STATE \textbf{return} $semantic\_mask$
\end{algorithmic}
\end{algorithm}

The large vision model-assisted co-inference method can significantly improve semantic segmentation accuracy by refining edge model-generated results. However, it faces two challenges: On one hand, it entails the uploading of samples to the cloud, potentially leading to increased latency. Therefore, the edge-cloud collaboration strategy is essential for solving this problem, as will be explained in section~\ref{3.3}. On the other hand, since the effectiveness of edge-cloud joint inference relies on the labels produced by the edge model, if the edge model encounters difficulties of environment mobility, thereby resulting in data drift and heterogeneity, the accuracy of large vision model-assisted co-inference will also diminish accordingly. Therefore, we need to adaptively update the edge model and the edge-cloud collaboration strategy, which will be elaborated in section~\ref{3.4}.

\subsection{Hard Input Mining Strategy}
\label{3.3}
The hard input mining strategy is pivotal in the LAECIPS architecture. Too many identified hard inputs will result in high inference latency while too few ones will lead to a decrease in the accuracy of handling corner cases. Existing methods rely on loss values~\cite{cvpr16} or confidence scores~\cite{ICLR18} to identify hard inputs. However, calculating loss values during inference is challenging due to unknown labels, and confidence score-based methods lack adaptability to changing environments. To address this, we propose a neural network-based hard input mining model, denoted as $h$. We adopt Resnet18~\cite{resnet} as the basic network structure of model $h$, whose input is the logit layer output of $f(x)$. This model determines if a data input for edge inference is a hard or easy input, represented as $h(f(x)) \in [0,1]$.
Based on this hard input mining model, the result of cloud-edge collaborative inference is:
\begin{equation}
(F, f, h)(x)=\left\{
\begin{aligned}
f(x) & \ \ \ if \ h(f(x)) > \delta\\
F(x) & \ \ otherwise. 
\end{aligned}
\right.
\end{equation}

We denote the loss of the model's output as
\begin{equation}
L(F, f, h, x, y)=\left\{
\begin{aligned}
l(f(x),y) & \ \ \ if \ h(f(x)) > \delta\\
l(F(x),y) & \ \ otherwise.
\end{aligned}
\right.
\end{equation}
and the inference latency of the model as 
\begin{equation}
\label{eq-delay}
delay(F, f, h, x)=\left\{
\begin{aligned}
d(f(x)) & \ \ \ if \ h(f(x)) > \delta\\
d(F(x)) & \ \ otherwise.
\end{aligned}
\right.
\end{equation}
In (\ref{eq-delay}), $d(F(x))$ includes the cloud model inference latency and the communication latency between cloud and edge.

Then, the loss of cloud-edge collaborative inference is:
\begin{equation}
\resizebox{\linewidth}{!}{$
    \begin{aligned}
        & \mathbb{E}_{P(x,y)}\mathbb{E}_{h(f(x))}[L(F, f, h, x, y)] = \\
         \mathbb{E}_{P(x,y)}[h(f(x)) & *l(f(x),y)  + (1-h(f(x)))*l(F(x),y)], 
    \end{aligned}
    $}
\end{equation}
and the latency of cloud-edge collaborative inference is:
\begin{equation}
\label{latency}
\resizebox{\linewidth}{!}{$
    \begin{aligned}
        & \mathbb{E}_{P(x,y)}\mathbb{E}_{h(f(x))}[delay(F, f, h, x)] = \\
\mathbb{E}_{P(x,y)}[h(f(x)) & *d(f(x)) + (1-h(f(x)))*d(F(x))].
    \end{aligned}
    $}
\end{equation}
We aim to improve inference accuracy while meeting the requirements of task process latency. Therefore, the overall optimization objective is:
\begin{equation}
\label{optimization-objective-1}
    \begin{aligned}
        \min_{F,f\in \mathbb{F}, h \in \mathbb{H}}{\mathbb{E}_{P(x,y)}\mathbb{E}_{h(f(x))}[L(F, f, h, x, y)]} \\
        s.t. \ \ \mathbb{E}_{P(x,y)}\mathbb{E}_{h(f(x))}[delay(F, f, h, x)] < delay_{max}\ .
    \end{aligned}
\end{equation}
Since the inference delay is independent of the inference inputs, we can simplify the inference delay as follows:
\begin{equation}
    \begin{aligned}
        d(f(x)) = d(f) = d_1 \\
        d(F(x)) = d(F) = d_0\ .
    \end{aligned}
\end{equation}
Thus, the cloud-edge collaborative inference delay in (\ref{latency}) can be simplified as:
\begin{equation}
    \begin{aligned}
        \mathbb{E}_{P(x,y)}\mathbb{E}_{h(f(x))}[delay(F, f, h, x)] = \\
        (d_1-d_0)*\mathbb{E}_{P(x,y)}[h(f(x))] + d_0\ .
    \end{aligned}
\end{equation}
Therefore, the constraint in (\ref{optimization-objective-1}) can be simplified as:
\begin{equation}
    \begin{aligned}
        \mathbb{E}_{P(x,y)}[h(f(x))] > \frac{d_0 -delay_{max}}{delay_{max} - d_1}\ .
    \end{aligned}
\end{equation}
Thus, the optimization objective in (\ref{optimization-objective-1}), which satisfies the KKT conditions~\cite{kkt}, can be rewritten as:
\begin{equation}
\label{optimization-objective-2}
\resizebox{\linewidth}{!}{$
    \min_{F,f\in \mathbb{F}, h \in \mathbb{H}}{\mathbb{E}_{P(x,y)}\mathbb{E}_{h(f(x))}[L(F, f, h, x, y)] + \beta * \mathbb{E}_{P(x,y)}[-log(h(f(x)))].}
    $}
\end{equation}

\subsection{Adaptive Update Process}
\label{3.4}
Since $F$ represents the large vision model-assisted inference function, there is no need to optimize $F$. Therefore, the optimization targets are $f$ and $h$. Additionally, since it is difficult to obtain the true label $y$ of a sample $x$ in a real environment, we optimize $f$ and $h$ using the large vision model-assisted inference result $F(x)$. The optimization objective in (\ref{optimization-objective-2}) can be further rewritten as:
\begin{equation}
\label{optimization-objective-3}
\resizebox{\linewidth}{!}{$
    \min_{f\in \mathbb{F}, h \in \mathbb{H}}{\mathbb{E}_{P(x,y)}[h(f(x))*l(f(x),F(x))] + \beta * \mathbb{E}_{P(x,y)}[-log(h(f(x)))].}
    $}
\end{equation}
The model update process can be divided into two steps: in the first step, we freeze $h$ and update $f$:
\begin{equation}
\label{f1loss}
    \begin{aligned}
        L_{f} = l(f(x),F(x)) \\
        \theta_{f} = \theta_{f} - \eta \bigtriangledown L_{f}\ .
    \end{aligned}
\end{equation}
Then, we freeze $f$ and update $h$:
\begin{equation}
\label{hloss}
    \begin{aligned}
        L_h  & = h(f(x))*l(f(x),F(x)) + \beta * -log(h(f(x))) \\
        \theta_{h} & = \theta_h - \eta \bigtriangledown L_{h}\ .
    \end{aligned}
\end{equation}

The overall workflow of LAECIPS combines the large vision model-assisted inference, hard input mining, and the adaptive update process, as shown in Algorithm~\ref{adaptive}.

\begin{algorithm}[tb]
\caption{Adaptive Edge-Cloud Collaboration algorithm}
\label{adaptive}
\textbf{Initialize}: Pretrained Edge Model: $\theta_{f}$, Cloud Model: $SAM$, Pretrained Hard Input Mining Model: $\theta_{h}$. \\
\textbf{Parameters}: confidence threshold: $threshold$, max replay buffer size: $maxsize$, max continual training interval: $maxtime$. \\
\textbf{Output}: Edge inference result: $edge\_result$, Co-inference result: $assisted\_result$.
\begin{algorithmic}[1] 
\STATE $replay\_buffer \gets \emptyset$
\FOR{$img \in samples$}
    \STATE Edge Node Collect $img$
    \STATE $edge\_result \gets f(img)$  
    \STATE \textcolor{blue}{$\backslash*$ perform hard input mining$*\backslash$}
    \STATE $confidence \gets h(edge\_result)$ 
    \IF{$confidence > threshold$}
        \STATE  \textbf{output} $edge\_result$
        \STATE Continue
    \ELSE
        \STATE Upload $img$ to Cloud Node
        \STATE \textcolor{blue}{$\backslash*$ perform large vision model assisted inference$*\backslash$}
        \STATE $mask \gets SAM(img)$
        \STATE Edge node Download $mask$ from Cloud Node
        \STATE $assisted\_result \gets Assisted\_Inference(edge\_result, mask)$ by Algorithm~\ref{Joint-Inference}
        \STATE \textbf{output} $assisted\_result$
        \STATE $replay\_buffer$ append $(img, assisted\_result)$
    \ENDIF
    \STATE \textcolor{blue}{$\backslash*$ perform adaptive update process$*\backslash$}
    \IF{$size(replay\_buffer) > maxsize$ or $timeinterval > maxtime$}
        \STATE Train $\theta_{f}$ with $replay\_buffer$ by Equation(\ref{f1loss})
        \STATE Train $\theta_{h}$ with $replay\_buffer$ by Equation(\ref{hloss})
        \STATE $replay\_buffer \gets \emptyset$
    \ENDIF
\ENDFOR
\end{algorithmic}
\end{algorithm}

\section{Theoretical Analysis of LAECIPS}
\label{sec:theory}

The system's generalization ability will greatly affect its actual effectiveness when deploying in a real-world dynamic IoT environment. In this section, we theoretically analyze the generalization boundary of the proposed system LAECIPS to prove its feasibility.

Based on the optimization objective from equation (\ref{optimization-objective-3}), the expected loss for the semantic segmentation function $f$ and hard input mining strategy $h$ is defined as follows:
\begin{equation}
\begin{aligned}
    R(f,h) & = \mathbb{E}_{P(x,y)}[h(f(x))*l(f(x),F(x))] \\
    & + \beta * \mathbb{E}_{P(x,y)}[-log(h(f(x)))] \ .
\end{aligned}
\end{equation}

\newtheorem{theorem}{Theorem}
\begin{theorem}
\label{generalization-bound}
Let $f$ be the family of semantic segmentation functions taking values in $[0,1]^{M \times H \times W}$, $h$ be the family of hard input mining functions taking values in $[0,1]$. We denote by $\widehat{R}_S(f,h)$ the empirical loss of function $(f,h)$ over the Sample $S$. Then, for any $\delta > 0$, with probability as least $1 - \delta$ over the draw of a sample $S$ of size $m$, the following holds for all $(f, h) \in \mathbb{F} \times \mathbb{H}$, where $\mathcal{R}_m$ represents the Rademacher complexity~\cite{Rm}:

\begin{equation}
	R(f,h) \le \widehat{R}_S(f,h) + (1 + \beta)\mathcal{R}_m(\mathbb{H}) + \mathcal{R}_m(\mathbb{F}) + \sqrt{\frac{log\frac{1}{\delta}}{2m}} \ .
\end{equation}
\end{theorem}

\begin{proof}
Let $l_{\mathbb{F},\mathbb{H}}$ be the family of functions $l_{\mathbb{F},\mathbb{H}} = \{(x,y) \rightarrow L(f, h, x, y), f \in \mathbb{F}, h \in \mathbb{H}\}$. By the general Rademacher complexity bound ~\cite{Rademacher}, with probability at least $1 - \delta$, the following holds for all $(f, h) \in \mathbb{F} \times \mathbb{H}$:

\begin{equation}    
	R(f, h) \le \widehat{R}_S(f, h) + 2\mathcal{R}_m(l_{\mathbb{F},\mathbb{H}}) + \sqrt{\frac{log\frac{1}{\delta}}{2m}}\ .
\end{equation}

Now, the Rademacher complexity can be bounded as follows:

\begin{equation}
\begin{aligned}
	& \mathcal{R}_m(l_{\mathbb{F},\mathbb{H}}) = \mathbb{E}_{\sigma}[\sup_{(f, h) \in \mathbb{F}  \times \mathbb{H}}\frac{1}{m}\sum_{i=1}^{m}\sigma_i * h(f(x_i)) \\
 & *l(f(x_i),F(x_i)) + \sigma_i * \beta * (-log(h(f(x_i))))] \\
	& \le \mathbb{E}_{\sigma}[\sup_{(f, h) \in \mathbb{F} \times \mathbb{H}}\frac{1}{m}\sum_{i=1}^{m}\sigma_i * h(f(x_i))*l(f(x_i),F(x_i))] \\
 & + \beta * \mathbb{E}_{\sigma}[\sup_{(f, h) \in \mathbb{F} \times \mathbb{H}}\frac{1}{m}\sum_{i=1}^{m}\sigma_i * (-log(h(f(x_i))))].
\end{aligned}
\end{equation}

\newtheorem{lemma}{Lemma}
\begin{lemma}
\label{sec:lemma}
Let $\mathbb{F}_1$ and $\mathbb{F}_2$ be two families of functions mapping $X$ to $[0, 1]$.
Let $\mathbb{F} = \{f_1*f_2: f_1 \in \mathbb{F}_1, f_2 \in \mathbb{F}_2\}$. Then, the empirical Rademacher complexities of $\mathbb{F}$ for any sample $S$ of size $m$ are bounded: 
\begin{equation}
    \widehat{\mathcal{R}_S}(\mathbb{F}) \le 2(\widehat{\mathcal{R}_S}(\mathbb{F}_1) + \widehat{\mathcal{R}_S}(\mathbb{F}_2))
\end{equation}
The proof of lemma~\ref{sec:lemma} could be found in~\cite{lemma1}

\end{lemma}

By lemma~\ref{sec:lemma}, the Rademacher complexity of products of indicator functions can be bounded by the sum of the Rademacher complexities of each indicator function class, thus:

\begin{equation}
\begin{aligned}
	& \mathbb{E}_{\sigma}[\sup_{(f, h) \in \mathbb{F} \times \mathbb{H}}\frac{1}{m}\sum_{i=1}^{m}\sigma_i * h(f(x_i))*l(f(x_i),F(x_i))] \\
	& \le \mathbb{E}_{\sigma}[\sup_{(f, h) \in \mathbb{F} \times \mathbb{H}}\frac{1}{m}\sum_{i=1}^{m}\sigma_i * h(f(x_i))] \\
 & + \mathbb{E}_{\sigma}[\sup_{f \in \mathbb{F} }\frac{1}{m}\sum_{i=1}^{m}\sigma_i * l(f(x_i),F(x_i))] \ .
\end{aligned}
\end{equation}

So, the Rademacher complexity can be bounded as follows:

\begin{equation}
\begin{aligned}
	 \mathcal{R}_m(l_{\mathbb{F},\mathbb{H}}) & \le (1 + \beta) * \mathbb{E}_{\sigma}[\sup_{(f, h) \in \mathbb{F} \times \mathbb{H}}\frac{1}{m}\sum_{i=1}^{m}\sigma_i * h(f(x_i))] \\
  & + \mathbb{E}_{\sigma}[\sup_{f \in \mathbb{F} }\frac{1}{m}\sum_{i=1}^{m}\sigma_i * l(f(x_i),F(x_i))] \\
	 & \le (1 + \beta) * \mathbb{E}_{\sigma}[\sup_{h \in \mathbb{H}}\frac{1}{m}\sum_{i=1}^{m}\sigma_i * h(f(x_i))] \\
  & + \mathbb{E}_{\sigma}[\sup_{f \in \mathbb{F} }\frac{1}{m}\sum_{i=1}^{m}\sigma_i * l(f(x_i),F(x_i))] \\
	 & \le (1 + \beta)\mathcal{R}_m(\mathbb{H}) + \mathcal{R}_m(\mathbb{F}) .
\end{aligned}
\end{equation}

\end{proof}

This theorem gives generalization guarantees for learning the semantic segmentation function $f$ and hard input mining function $h$ that admit Rademacher complexities in $O(\frac{1}{\sqrt{m}})$.

Theorem~\ref{generalization-bound} indicates that the maximum generalization error of LAECIPS is bounded provided that the maximum generalization error of the semantic segmentation models and hard input mining strategy deployed in LAECIPS is controllable.  
Therefore, it is theoretically feasible to deploy large visual models, small models, and hard input mining strategies in the LAECIPS framework for co-inference in a plug-and-play manner.

\section{Experiments}
\label{sec:evaluation}


\subsection{Experiment Setup}

\paragraph{Hardware and Software Systems}

\begin{figure}[t]
\centering
\includegraphics[width=1.0\columnwidth]{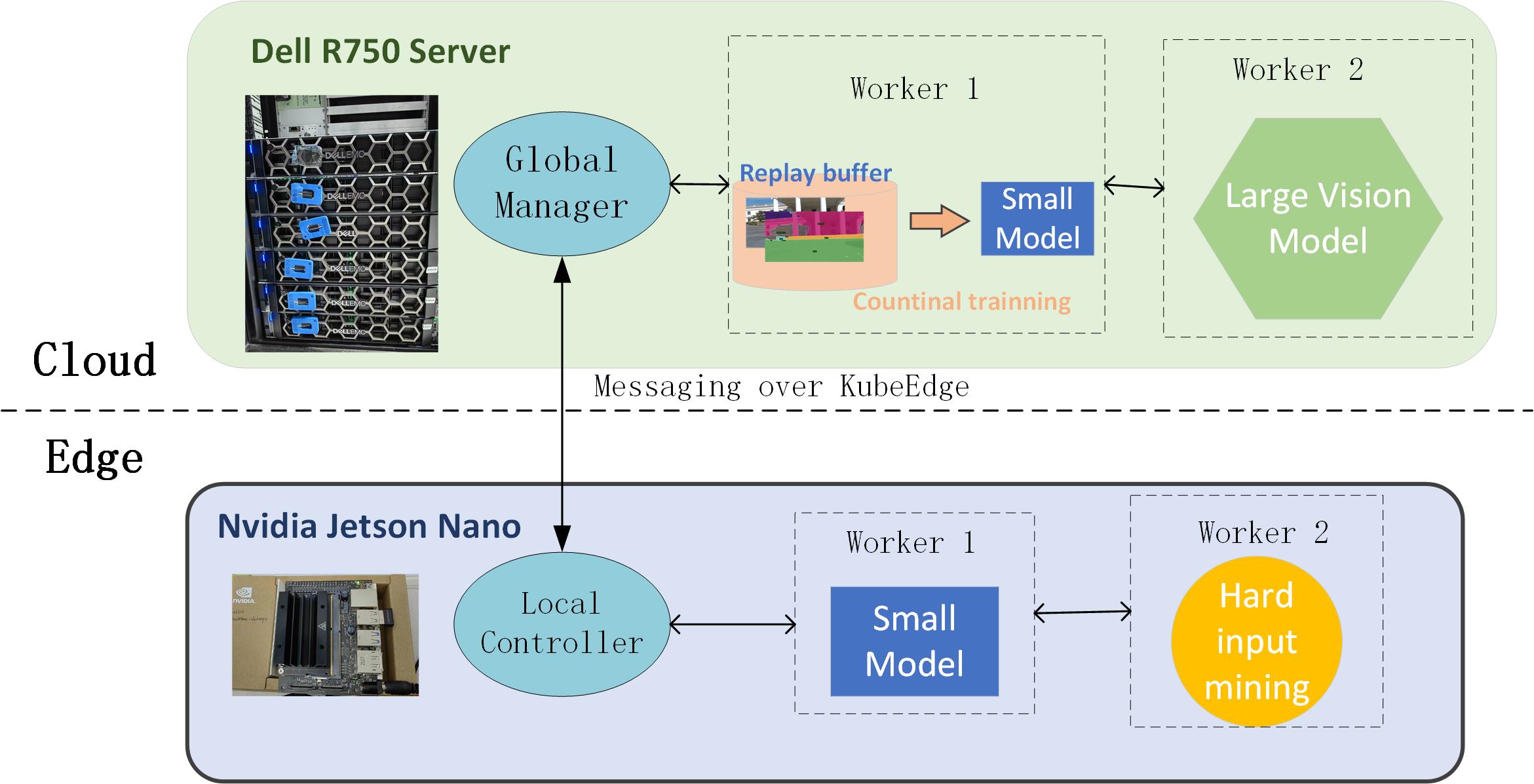}
\caption{Experimental setup for LAECIPS}
\label{setup}
\end{figure}

We implemented a system prototype of the proposed LAECCIPS framework for real-world robotic semantic segmentation and conducted experiments on it for performance evaluations.
In the hardware setup, we use the Nvidia Jetson Nano~\cite{nano}, which is commonly used in real-world robotic devices, as the edge node. For the cloud node, we have a Dell R750 server with a 48-core Intel Xeon Silver 4310 CPU @ 2.10GHz, 256GB of memory, and 2 Nvidia GeForce 3090 GPUs. The cloud node and edge node are connected via WLAN with a network bandwidth of 4Mbps. We've implemented LAECIPS using the distributed AI testing framework Ianvs~\cite{kubeedge} based on Kubeedge, deploying small models on the Jetson Nano and the large vision model on the Dell R750 server as shown in Figure~\ref{setup}. 

\paragraph{Datasets}
Semantic segmentation is a typical task in the IoT embodied intelligence system and also a fundamental task in the fields of robotics and autonomous driving. To validate the effectiveness of our proposed LAECIPS in the real-world IoT perception environment, we selected four typical real-world semantic segmentation datasets:
\begin{itemize}
    \item The Cloud-Robotics dataset~\cite{cloud-robotics} contains 2600 semantic segmentation images collected by intelligent robotic dogs in the Shenzhen Industrial zone, mainly applicable to robot scenes in semi-enclosed areas.
    \item The Cityscapes dataset~\cite{cityscapes} contains 5000 semantic segmentation images collected by smart cars in multiple cities in Germany, mainly applicable to autonomous driving scenes in open-world environments.
    \item The ADE20K dataset~\cite{zhou2017scene} contains 20,000 semantic segmentation images, covering various scenes from indoor to outdoor, natural to urban, and can be used for tasks like scene understanding and image segmentation in robotics and autonomous driving.
    \item The SYNTHIA dataset~\cite{Ros_2016_CVPR} contains 9,000 semantic segmentation images, consisting of photo-realistic frames rendered from a virtual city and includes precise pixel-level semantic annotations.
\end{itemize}

\begin{figure}[t]
\centering
\includegraphics[width=1.0\columnwidth]{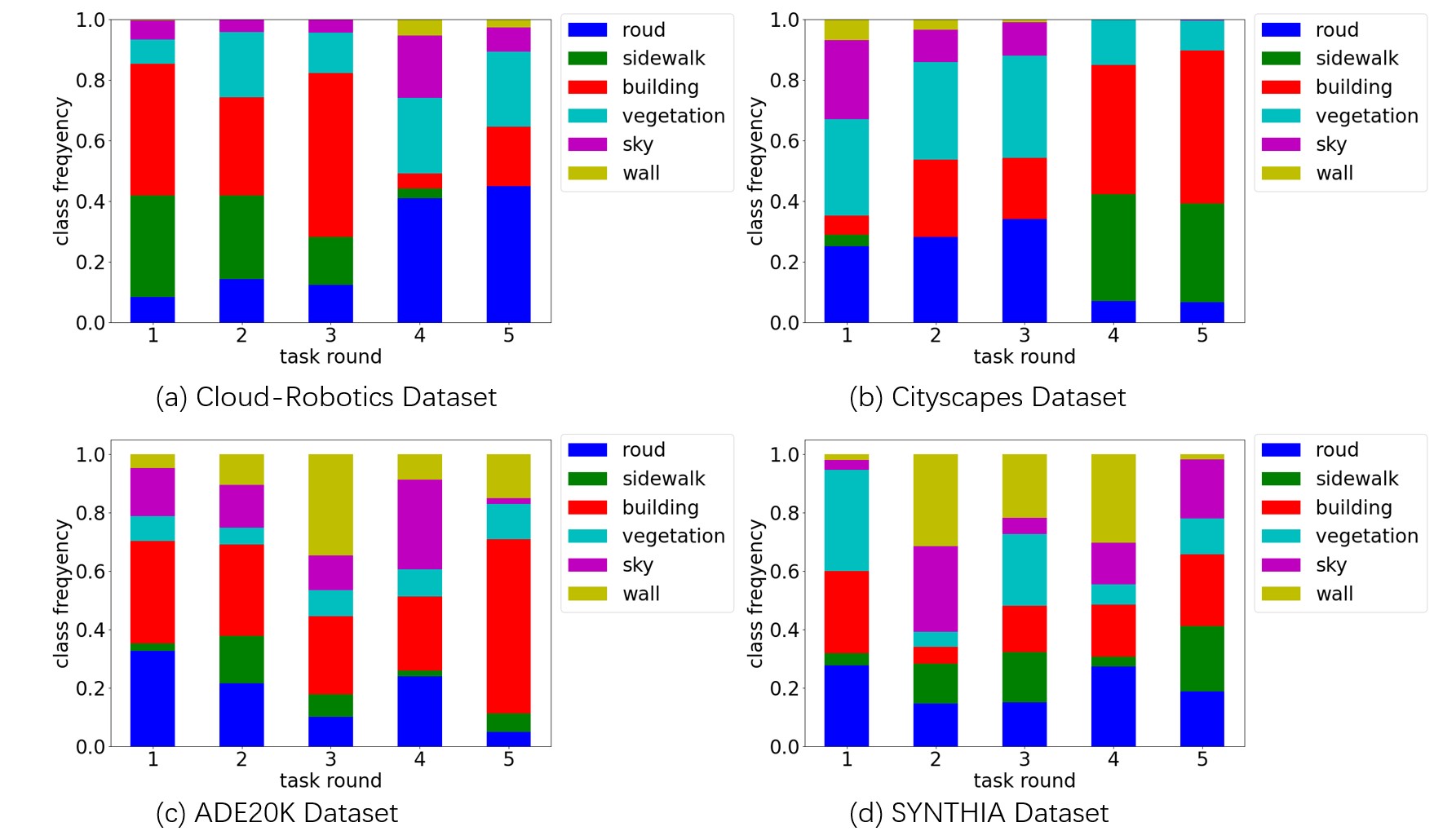} 
\caption{Change of class distribution in different tasks}
\label{freq}
\end{figure}

To test the system performance in a dynamic IoT environment in our experiments, we divide the datasets into 5 tasks in chronological order. As shown in Figure~\ref{freq}, the class distributions in the datasets all vary across different tasks due to the change in the task locations, e.g., from indoor to outdoor and from playground to road. These tasks with different class distributions reflect common data drifting phenomena in real-world applications; therefore, testing the system performance across different tasks allows us to evaluate the adaptability of the proposed method to dynamic environments.

\paragraph{Compared Methods}

We first compared three different baseline frameworks:
\begin{itemize}
    \item CLOUD: Upload all inputs to the cloud node for processing by the large visual model.
    \item EDGE: Process all inputs on the edge node using the small model. 
    \item DCSB: DCSB~\cite{cao2023edge} is the SOTA method for big/little model cooperation in object detection tasks. We implement DCSB in the semantic segmentation task with the same hard input mining strategy of LAECIPS for a fair comparison. The difference between this framework and our proposed LAECIPS framework is that DCSB does not dynamically update the small model.
\end{itemize} 

Besides that, we also employed three typical hard input mining strategies, MESS~\cite{ECCV22}, SM~\cite{CODES15}, and SPP~\cite{ICLR18}, to evaluate the effectiveness and generalization of our LAECIPS method.

\begin{itemize}

\item MESS is the SOTA method proposed for early exit semantic segmentation, which could also be used in hard input mining. It calculates the confidence score of an inference result by counting the proportion of pixels with a maximum probability distribution greater than a certain threshold.

\item SM is the classic method used in edge-cloud collaboration. It calculates the confidence score based on the difference between the maximum probability distribution and the second maximum probability distribution in the inference result.

\item SPP is the baseline method for hard input mining. It calculates the confidence score based on the maximum probability distribution in the inference result.

\end{itemize}

For a fair comparison, the above three algorithms for hard input mining will be applied in the proposed framework in the semantic segmentation task during the experimental process.

\paragraph{Evaluation Metrics}
The metrics we test in the experiment include mIoU, Cloud Upload Rate (CUR), and latency. mIoU measures the model's inference accuracy in semantic segmentation tasks. CUR represents the proportion of images uploaded to the cloud, reflecting the communication overhead of edge-cloud co-inference. The latency is the average time for completing the co-inference process for image inputs. 

The calculation of the inference mIoU accuracy is as follows:
\begin{equation}
\begin{aligned}
	mIoU(F,f,h) & = \frac{1}{N}\sum_{i=1}^N[\mathds{1}(h(f(x_i))\ge \delta)*IoU(f(x_i), y) \\
	& + \mathds{1}(h(f(x)) < \delta) * IoU(F(x_i), y)].
\end{aligned}
\end{equation}

The calculation of the Cloud Upload Rate(CUR) is as follows:
\begin{equation}
	CUR = \frac{1}{N}\sum_{i=1}^N\mathds{1}(h(f(x_i)) < \delta).
\end{equation}

The calculation of the latency is as follows:
\begin{equation}
    latency = \frac{1}{N}\sum_{i=1}^N(delay(x_i)).
\end{equation}

\paragraph{Base Models}
We adopted the SAM model for the large vision model deployed in the cloud. We employed the RFNet model~\cite{rfnet} for the small model, which is more suitable for deployment on edge devices because of its lightweight structure. The Hard input mining model adopts a Resnet18-based model structure with less than 20M parameters, which will not cause excessive burden on the edge nodes. In the experiments, the SAM model uses the pre-trained parameters published by Meta without additional training. the RFNet model for hard input mining is pre-trained on the training set of the datasets, in which the samples $x$ that satisfy $mIoU(F(x)) - mIoU(f(x)) \ge 0.1$ are taken as hard inputs and the other samples are considered as easy inputs. The learning rate and epochs for training the RFNet model are set to 0.001 and 50.

\subsection{Experimental Results}

\begin{figure}[t]
\centering
\includegraphics[width=1.0\columnwidth]{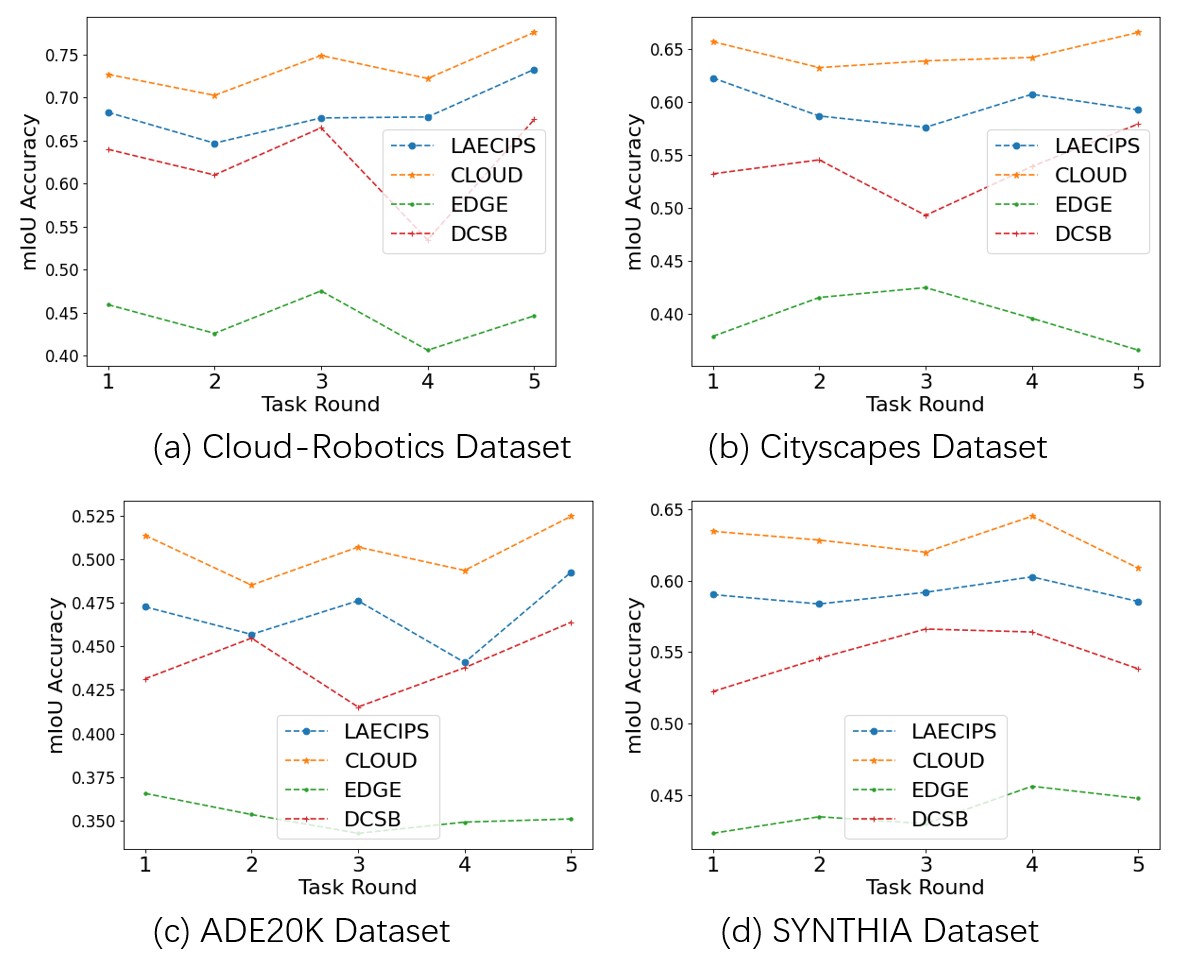} 
\caption{Effectiveness of LAECIPS in Different Tasks.}
\label{ablation}
\end{figure}

\begin{figure}[t]
\centering
\includegraphics[width=1.0\columnwidth]{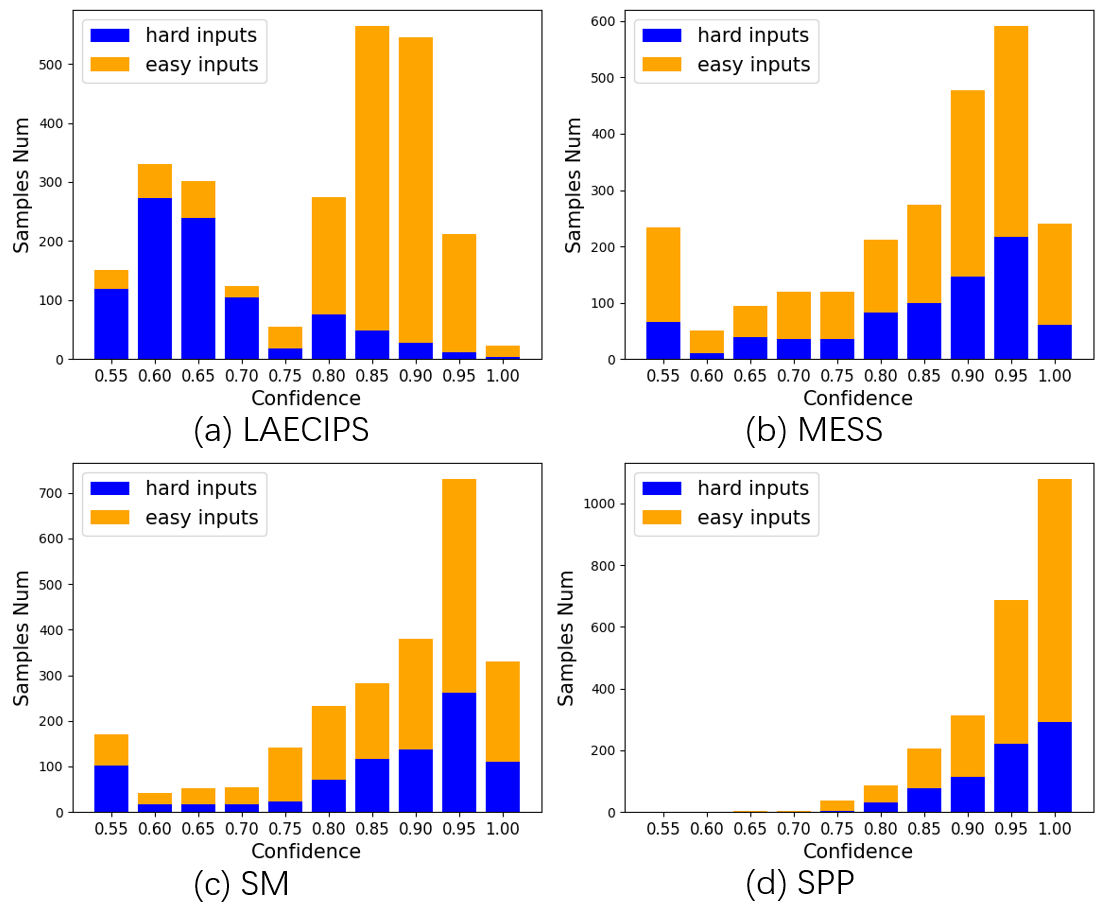} 
\caption{Hard input and easy input histogram of different algorithms in Cloud-Robotics Dataset}
\label{hard-example-result}
\end{figure}

\begin{figure}[t]
\centering
\includegraphics[width=1.0\columnwidth]{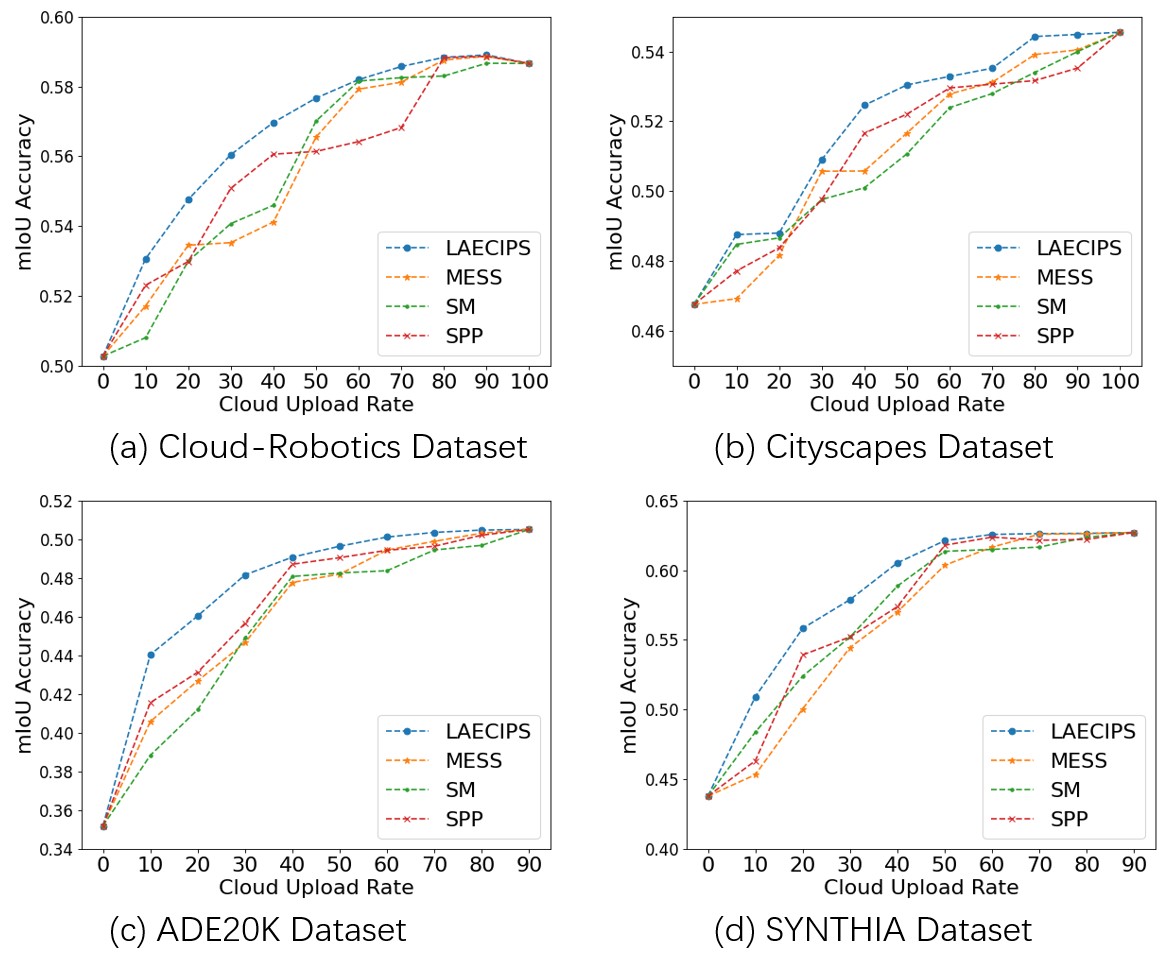} 
\caption{mIoU under different CUR}
\label{Comparision of Different Cloud Update Rate}
\end{figure}

\begin{figure*}[t]
\centering
\includegraphics[width=2.0\columnwidth]{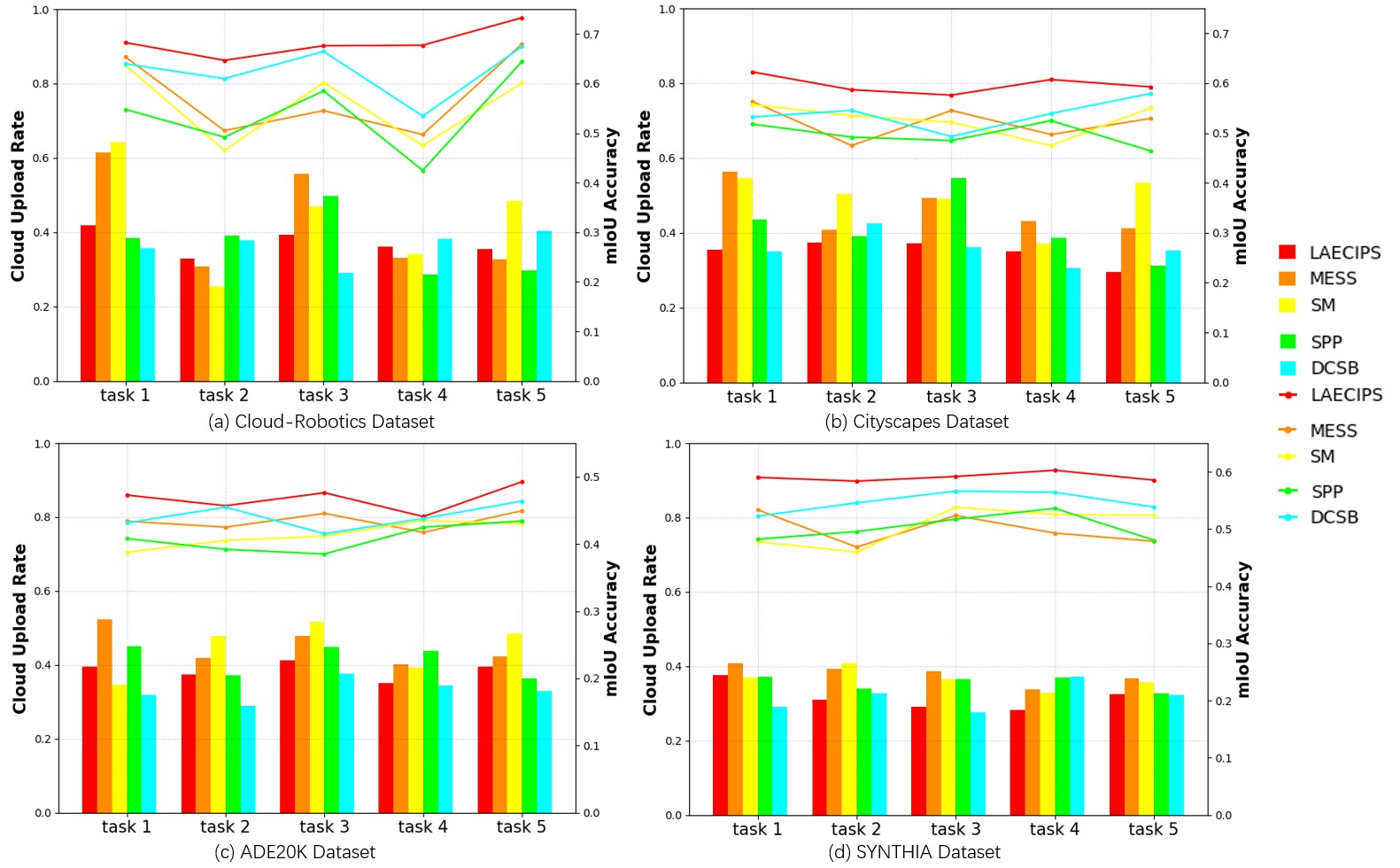} 
\caption{mIoU and CUR under different tasks}
\label{Comparison of Different Algorithms}
\end{figure*}

Tables~\ref{table1} and \ref{table2}, and Figures~\ref{ablation}, \ref{hard-example-result}, \ref{Comparision of Different Cloud Update Rate} and \ref{Comparison of Different Algorithms} show the experimental results of LAECIPS and other frameworks, algorithms in different datasets. Through these experimental results, we aim to answer the following research questions.
\begin{description}
    \item[Q1.] {\em How effective is the edge-cloud collaboration in our proposed LAECIPS framework?}
\end{description}

\begin{table}[htbp]
\caption{Comparision of average mIoU, CUR, and inference latency for edge-cloud collaboration of LAECIPS}
\label{table1}
\centering
\resizebox{\columnwidth}{!}{
\begin{tabular}{|c|c|c|c|c|}
    \hline
    \textbf{Dataset} & \textbf{Method} & \textbf{mIoU} & \textbf{CUR} & \textbf{latency} \\
    \hline
    Cloud-Robotics & LAECIPS & 0.683 & 37.12\% & 2.60 \\
    & Cloud & 0.735 & 100\% & 5.11 \\
    & Edge & 0.442 & 0\% & 1.12 \\
    & DCSB & 0.624 & 36.22\% & 2.56 \\
    \hline
     Cityscapes & LAECIPS & 0.597 & 34.98\% & 2.74 \\
    & Cloud & 0.647 & 100\% & 5.83 \\
    & Edge & 0.396 & 0\% & 1.09 \\
     & DCSB & 0.537 & 35.86\% & 2.79 \\
    \hline
    ADE20K & LAECIPS & 0.467 & 38.52\% & 2.52 \\
    & Cloud & 0.504 & 100\% & 4.88 \\
    & Edge & 0.352 & 0\% & 1.05 \\
    & DCSB & 0.441 & 33.12\% & 2.32 \\
    \hline
    SYNTHIA & LAECIPS & 0.591 & 31.71\% & 2.33 \\
    & Cloud & 0.627 & 100\% & 5.07 \\
    & Edge & 0.438 & 0\% & 1.06 \\
    & DCSB & 0.547 & 31.81\% & 2.34 \\
    \hline 
\end{tabular}}
\end{table}

To answer this question, we make two observations from the experimental results presented in Figure~\ref{ablation} and Table~\ref{table1}. Firstly. Compared with the EDGE framework which handles all the input at the edge, the LAECIPS framework uses the large vision model deployed at the cloud to process hard inputs that the edge small model cannot handle well, and thus improves mIoU accuracy on different datasets by 10\% to 20\%. Compared with the DCSB framework, LAECIPS uses the hard inputs collected from the cloud to continuously train the edge small model, resulting in a 3\% to 6\% improvement in mIoU accuracy. Compared with the CLOUD framework, which has the highest accuracy at the cost of delay and communication overheads, LAECIPS only sacrifices a marginal degradation (2\% to 5\%) in mIoU accuracy. Those results demonstrate that the LAECIPS method can effectively improve the model's inference accuracy.

Secondly. Table~\ref{table1} shows the average inference latency and CURs. Compared to the CLOUD framework, LAECIPS saves over 60\% of inference time and communication overhead by handling most of the easy inputs at the edge. Compared to the current SOTA DCSB framework, LAECIPS has very similar inference latency and communication overhead since they use the same hard input mining strategy. This proves that the LAECIPS can effectively reduce inference latency and communication overhead.

\begin{description}
    \item[Q2.] {\em Is our cloud-edge collaboration method more effective in identifying hard inputs compared to other hard input mining algorithms?}
\end{description}

\begin{table*}[htbp]
\caption{Average mIoU, CUR, and inference latency of different tasks.}
\label{table2}
\centering
\resizebox{2.0\columnwidth}{!}{
\begin{tabular}{|c|c|c|c|c|c|c|c|c|c|c|c||c|c|c|}
    \hline
    \textbf{Dataset} & \textbf{Method} & \textbf{1-mIoU} & \textbf{1-CUR} & \textbf{2-mIoU} & \textbf{2-CUR} & \textbf{3-mIoU} & \textbf{3-CUR} & \textbf{4-mIoU} & \textbf{4-CUR} & \textbf{5-mIoU} & \textbf{5-CUR} & \textbf{avg mIoU} & \textbf{avg CUR} & \textbf{avg latency} \\
    \hline
    Cloud-Robotics & LAECIPS & \textbf{0.682} & 41.94\% & \textbf{0.647} & 32.85\% & \textbf{0.676} & 39.29\% & \textbf{0.677} & 36.04\% & \textbf{0.733} & 35.55\% & \textbf{0.683} & 37.12\% & 2.60 \\
    & DCSB & 0.639 & \textbf{35.70\%} & 0.610 & 37.83\% & 0.664 & \textbf{29.15\%} & 0.534 & 38.17\% & 0.674 & 40.35\% & 0.624 & \textbf{36.22\%} & \textbf{2.56} \\
    & MESS & 0.654 & 61.38\% & 0.505 & 30.83\% & 0.545 & 55.76\% & 0.497 & 33.14\% & 0.679 & 32.72\% & 0.576 & 42.93\% & 2.83 \\
    & SM & 0.637 & 64.17\% & 0.465 & \textbf{25.41\%} & 0.602 & 47.08\% & 0.475 & 34.23\% & 0.601 & 48.45\% & 0.556 & 43.86\% & 2.87 \\
    & SPP & 0.548 & 38.5\% & 0.492 & 39.1\% & 0.585 & 49.75\% & 0.425 & \textbf{28.72\%} & 0.644 & \textbf{29.77\%} & 0.538 & 37.86\% & 2.63 \\
    \hline
     Cityscapes & LAECIPS & \textbf{0.623} & 35.45\% & \textbf{0.587} & \textbf{37.37\%} & \textbf{0.576} & 37.29\% & \textbf{0.607} & 35.05\% & \textbf{0.593} & \textbf{29.46\%} & \textbf{0.597} & \textbf{34.98\%} & \textbf{2.74} \\
     & DCSB & 0.532 & \textbf{34.99\%} & 0.546 & 42.47\% & 0.493 & \textbf{36.04\%} & 0.539 & \textbf{30.58\%} & 0.579 & 35.23\% & 0.537 & 35.86\% & 2.79 \\
    & MESS & 0.564 & 56.38\% & 0.475 & 40.83\% & 0.545 & 49.26\% & 0.497 & 43.14\% & 0.529 & 41.22\% & 0.522 & 46.33\% & 3.28 \\
    & SM & 0.557 & 54.67\% & 0.535 & 50.41\% & 0.522 & 49.08\% & 0.475 & 37.23\% & 0.551 & 53.45\% & 0.528 & 49.76\% & 3.44 \\
    & SPP & 0.518 & 43.5\% & 0.492 & 39.1\% & 0.485 & 54.75\% & 0.525 & 38.72\% & 0.464 & 31.27\% & 0.496 & 41.46\% & 3.05 \\
    \hline
    ADE20K & LAECIPS & \textbf{0.473} & 39.45\% & \textbf{0.457} & 37.37\% & \textbf{0.476} & 41.29\% & \textbf{0.441} & 35.05\% & \textbf{0.493} & 39.46\% & \textbf{0.468} & 38.52\% & 2.50 \\
    & DCSB & 0.431 & \textbf{31.84\%} & 0.455 & \textbf{28.84\%} & 0.415 & \textbf{37.7\%} & 0.438 & \textbf{34.37\%} & 0.464 & \textbf{32.87\%} & 0.441 & \textbf{33.12\%} & \textbf{2.32} \\
    & MESS & 0.434 & 52.38\% & 0.425 & 41.83\% & 0.445 & 47.76\% & 0.417 & 40.14\% & 0.449 & 42.22\% & 0.434 & 44.86\% & 2.77 \\
    & SM & 0.387 & 34.67\% & 0.405 & 47.91\% & 0.412 & 51.58\% & 0.435 & 39.23\% & 0.43 & 48.45\% & 0.414 & 44.36\% & 2.75 \\
    & SPP & 0.408 & 45.0\% & 0.392 & 37.1\% & 0.385 & 44.75\% & 0.425 & 43.72\% & 0.434 & 36.27\% & 0.408 & 41.37\% & 2.64 \\
    \hline
     SYNTHIA & LAECIPS & \textbf{0.59} & 37.72\% & \textbf{0.584} & \textbf{30.96\%} & \textbf{0.592} & 29.02\% & \textbf{0.603} & \textbf{28.3\%} & \textbf{0.585} & 32.54\% & \textbf{0.591} & \textbf{31.71\%} & \textbf{2.33} \\
     & DCSB & 0.523 & \textbf{29.07\%} & 0.546 & 32.83\% & 0.566 & \textbf{27.69\%} & 0.564 & 37.16\% & 0.538 & \textbf{32.33\%} & 0.547 & 31.81\% & 2.34 \\
    & MESS & 0.534 & 40.71\% & 0.468 & 39.33\% & 0.523 & 38.68\% & 0.493 & 33.75\% & 0.478 & 36.82\% & 0.499 & 37.86\% & 2.58 \\
    & SM & 0.478 & 36.91\% & 0.46 & 40.9\% & 0.538 & 36.52\% & 0.525 & 33.01\% & 0.524 & 35.71\% & 0.505 & 36.61\% & 2.52 \\
    & SPP & 0.482 & 37.12\% & 0.495 & 34.05\% & 0.517 & 36.51\% & 0.536 & 37.02\% & 0.48 & 32.8\% & 0.502 & 35.49\% & 2.48 \\
    \hline
\end{tabular}}
\end{table*}

We answer this question by making two observations from Figure~\ref{hard-example-result} and Figure~\ref{Comparision of Different Cloud Update Rate}. Firstly, we classify the samples $x$ that satisfy the condition $mIoU(F(x)) - mIoU(f(x)) \ge 0.1$ as hard inputs. Fig.~\ref{hard-example-result} shows the differentiation between hard inputs and easy inputs based on the confidence scores of different algorithms. The MESS, SM and SPP methods simply use the maximum probability distribution of the inference result to calculate the confidence. Therefore, in complex semantic segmentation tasks, hard inputs and easy inputs cannot be distinguished directly according to their confidence scores. LAECIPS adopts a hard input mining strategy based on a neural network model, which can use all logit information of inference results to make the judgment; therefore it is more effective in discriminating hard and easy inputs. 

Secondly, as shown in Figure~\ref{Comparision of Different Cloud Update Rate}, we tested the inference mIoU accuracy under different CURs by adjusting the threshold $\delta$ with the same edge model. It can be seen that the inference accuracy of LAECIPS is higher than that of other methods under different CURs. The results indicate that LAECIPS introduces less amount of communication overhead compared to other methods for achieving the same level of inference accuracy, further validating the effectiveness of the LAECIPS method in identifying hard inputs.

\begin{description}
    \item[Q3.] {\em Is the LAECIPS algorithm more adaptable to dynamic environmental changes?}
\end{description}

We make two observations from Figure~\ref{Comparison of Different Algorithms} and Table~\ref{table2} to answer this question. Firstly, Figure~\ref{Comparison of Different Algorithms} shows the inference mIoU accuracy and CURs of various algorithms in different tasks. The class distributions of different tasks from the same dataset are significantly different as shown in Figure~\ref{freq}, which have certain impacts on the effectiveness of the edge small model and hard input mining algorithms, leading to fluctuations in the model's inference accuracy and CURs across different tasks. Therefore, the performance variances of the evaluated methods for handling different tasks reflect their adaptability to dynamic environments. 

The DCSB, MESS, SM, and SP methods all suffer large fluctuations in mIoU and CUR under different tasks due to their inability to dynamically update the edge small model. LAECIPS can adaptively adjust the edge small model according to the current task, so it can adapt to the dynamic environment and maintain stable performance, which enables LAECIPS to outperform other algorithms in various tasks across the 4 datasets in the experiments. 
Secondly, Table~\ref{table2} shows the mIoU, CUR, and inference latency under different tasks. Across different tasks, LAECIPS has the highest mIoU, further highlighting the effectiveness of the adaptive update process used in the LAECIPS framework. Since the hard input mining strategy of LAECIPS is more efficient, its CUR and inference delay are much lower than those of the MESS, SM, and SPP methods. Because DCSB adopts the same hard input mining strategy as LAECIPS but does not dynamically update the edge small model, it may achieve lower CUR and latency than LAECIPS in some datasets but at the price of lower mIoU due to the lack of adaptability to dynamic environments.

\section{Conclusion}
\label{sec: conclusion}
This paper delves into the new problem of online cloud-edge collaborative training and inference in dynamic environments, underscored by large vision models in the IoT perception landscape. The crux of this problem lies in discerning optimal collaboration strategies that cater to the real-time demands of edge sensing and computing while bolstering inference accuracy. 
Our solution, the LAECIPS framework, decouples its primary constituents -- a large vision model hosted on the cloud and a small model deployed at the edge -- and employs a hard input mining-based co-inference strategy to optimize their collaboration. 
With LAECIPS, only the hard inputs are deferred to the cloud, and the edge model is adaptively updated, learning from the pre-trained large vision model outputs to ensure resilience to dynamic environmental shifts. 
The generalization error bound of LAECIPS has been derived, and comprehensive evaluations on real-world robotic semantic segmentation benchmarks have been conducted.
Both theoretical and empirical results substantiate the viability and effectiveness of our proposed framework.
We believe that our work lays a solid foundation for large vision model-assisted edge-cloud collaboration and facilitates the development of IoT Embodied Intelligence systems.

Our work contributes a scalable, practical approach to implementing embodied intelligence for smart manufacturing, with immediate relevance to visual inspection, quality control, and beyond. Future research will explore the extension of LAECIPS to other embodied intelligence scenarios in smart manufacturing, including multi-robot collaboration, digital twin-enhanced decision making, and advanced human-machine teaming on the shop floor.

\bibliographystyle{cas-model2-names}

\bibliography{cas-refs}





\end{document}